\documentclass[10pt,twocolumn,letterpaper]{article}

\usepackage{iccv}
\usepackage{times}
\usepackage{epsfig}
\usepackage{graphicx}
\usepackage{amsmath}
\usepackage{amssymb}
\usepackage{subcaption}
\usepackage{float}

\usepackage{algorithm}
\usepackage{algpseudocode}
\let\oldemptyset\emptyset

\usepackage[pagebackref=true,breaklinks=true,letterpaper=true,colorlinks,bookmarks=false]{hyperref}

\iccvfinalcopy 

\def\iccvPaperID{5343} 

\ificcvfinal\fi
\begin{document}

\title{Deep Neural Network for Semantic-based Text Recognition in Images}


\author{Yi Zheng, Qitong Wang, Margrit Betke \\
Boston University\\
{\tt\small \{yizheng,wqt1996,betke\}@bu.edu}
\and
}

\maketitle

\begin{abstract}
   State-of-the-art text spotting systems typically aim to detect isolated words or word-by-word text in images of natural scenes and ignore the semantic coherence within a region of text.
   However, when interpreted together, seemingly isolated words may be easier to recognize. 
   On this basis, we propose a novel ``semantic-based text recognition'' (STR) 
   deep learning model that reads text in images with the help of understanding context. STR consists of several modules. We introduce the Text Grouping and Arranging (TGA) algorithm to 
   connect and order isolated text regions. 
   A text-recognition network interprets isolated words.
   Benefiting from semantic information, a sequence-to-sequence network model efficiently corrects inaccurate and uncertain phrases produced earlier in the STR pipeline.  We present experiments on two new distinct datasets that contain scanned catalog images of interior designs and photographs of protesters with hand-written signs, respectively. 
   Our results show that our STR model outperforms a baseline method that uses state-of-the-art single-word-recognition techniques on both datasets.  STR yields a high accuracy rate of 90\% on the catalog images and 71\% on the more difficult protest images, suggesting its generality in recognizing text.
   \vspace*{-0.5cm}
\end{abstract}

\section{Introduction}
    Recognizing text in images is a research problem that has attracted significant interest in the last few years due to its numerous potential applications in document image analysis, image retrieval, scene understanding, visual assistance, and so on.   Early work focused on images of printed text, which can be interpreted with traditional sliding-window-based Optical Character Recognition (OCR) techniques.
    In the last few years, great progress has been made with the development of methods that use deep convolutional neural networks (CNNs)~\cite{Jaderberg2016,ShiBaYa15} to ``spot text'' in natural scene images, that is, to both locate text target regions and then recognize the words in these regions.  
    
    \begin{figure}[t]
\begin{subfigure}{0.4\linewidth}
        \includegraphics[width=1.6in]{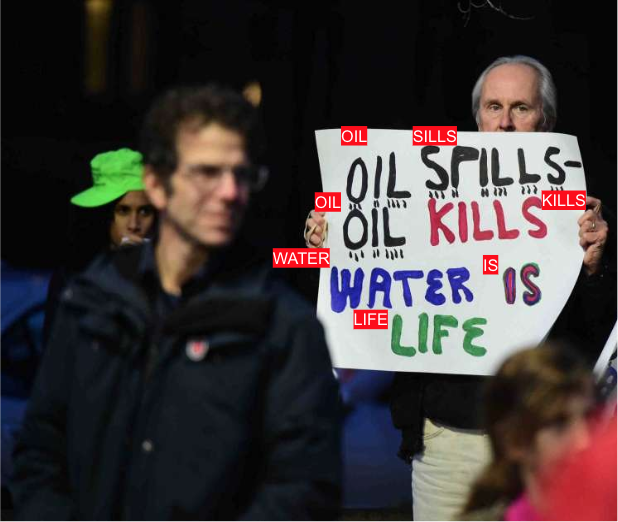}
        \caption{}
        \label{(a)}
    \end{subfigure}
    \hspace{2em}
    \begin{subfigure}{0.4\linewidth}
        \includegraphics[width=1.6in]{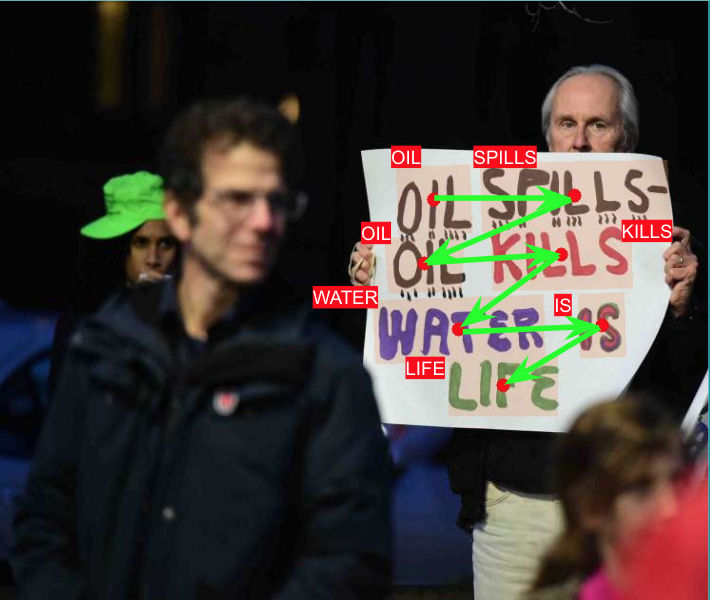}
        \caption{}
        \label{(b)}
    \end{subfigure}
    \vspace*{-0.2cm}
    \caption{In this work, we show that recognizing words in images may fail when they are considered 
    individual word by individual word, but including an analysis of semantic information can help correcting inaccurate predictions.  State-of-the-art text recognition misses the letter ``P'' and  misinterprets the word
    ``SPILLS''  as ``SILLS'' (Fig~\ref{(a)}), while our proposed
    semantic-based text recognition deep learning model STR groups and arranges the images of words, utilizes semantic information, and improves text recognition -- here, the bi-gram ``OIL SPILLS''  (Fig~\ref{(b)})}
\label{fig:task}
\vspace*{-0.3cm}
\end{figure}

     Although tremendous efforts have been devoted to improving the performance of text spotting models, being able to read text automatically in arbitrary images is still extremely challenging and remains an open problem.  This even applies to the seemingly ``easy'' domain of document images when they include photographs of natural scenes with text overlays.  We collected such images and offer a new large dataset with ground-truth labels, which we make publicly available. We scanned almost 8,000 pages of historic interior design catalogues.  The pages show photos of home interiors and furniture, text descriptions that include hundreds of thousands of words, and prices.  Analysis of this data is difficult due to the placement of the text in the photos and scanning artifacts.  
     
     To create a true research challenge for text spotting in outdoor scenes, we designed a second dataset that includes highly variable text, which appears on hand-written signs that people carry in street protests (Fig.~\ref{fig:task}).  Words may be geometrically distorted, and word phrases not horizontally aligned.  Furthermore, font type and size of hand-written text are highly irregular. 
   
     The ground-truth annotations that we provide in both of our datasets include group labels for coherent word phrases (e.g. a protest sign slogan) and paragraphs.  To the best of our knowledge, this property is unique.  Existing datasets do not provide such groupings. 
     They contain outdoor scenes with single-word text such as traffic signs (STOP), street names, or signs on building facades (restaurant names).    
     State-of-the-art methods that have been trained on existing datasets therefore treat every occurrence of text in an image as an isolated region that needs to be interpreted individually.  
     
     The innovative insight that our paper offers is that images with word groups contain semantic information that should not be ignored.  Semantic information should contribute to the ability of a model to read text.  In this paper, we show how a deep learning model can be designed and trained to take advantage of semantic information in order to recognize multi-word text in images.
     
    In this paper, we propose a novel semantic-based text recognition deep learning model, called \textbf{STR} (Fig.~\ref{fig:task}).  With this model, we make two main contributions:  Firstly, we put forward a layout analysis algorithm,  called \textbf{TGA}, with which the text regions that collectively make up a sentence or paragraph can be grouped together and arranged in the correct order. Secondly, we propose a Sequence-to-Sequence Learning Model~\cite{ShiBaYa15} that converts sequences from one domain to sequences in another domain, in order to correct the predicted results from single-word text recognition, improving the accuracy of the model by including semantic information. 
    
    The motivations of the sequence-to-sequence semantic-based spelling correction model are three fold. Firstly, the sequence-to-sequence model is used to translate a data sequence into another data sequence and widely used in the fields of  machine translation, audio recognition, etc.  Based on this, we treat predicted results from text recognition as input sequences and treat corrected results as output sequences to reach our spelling correction goal. Secondly, in order to take advantage of domain knowledge, the inputs to our model are phrases of words rather than single words, so that the model can combine domain knowledge and context to correct the result of our text recognition module. Thirdly, for the purpose of making the model to more effective in combining ``front and back memory information''  when correcting the spelling of words, we implemented a bidirectional Long Short-Term Memory (LSTM) network~\cite{schuster1997bidirectional} in the encoder part of our sequence-to-sequence model.
    
    To show the effectiveness of our proposed model, we conduct experiments on the aforementioned interior design and protest image datasets.
    
    \vspace{0.1cm}
    \noindent
    The main contributions of this paper are as follows:
    \vspace*{-0.2cm}
    \begin{itemize} 
    \item We introduce TGA, an effective algorithm that groups word regions in a phrase, sentence, or paragraph, and arranges them in the correct order.
    \item We use a sequence-to-sequence model to effectively correct the results from a text-recognition network based on semantic information.
    \item We create two new labeled datasets.
    \item Our STR model performs well in different experiments with different semantic contexts, suggesting its generality in recognizing text.
    \end{itemize}
    
    \begin{figure*}
    \begin{center}
        \includegraphics[width=0.9\linewidth]{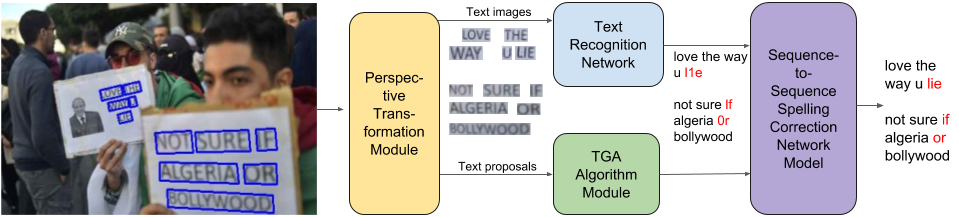}
    \end{center}
    \caption{Overall architecture of STR: First, a Perspective Transformation module produces cropped and normalized text images based on text bounding boxes. Text regions are grouped and arranged by the TGA algorithm module. The text recognition network predicts isolated words. Based on semantic information, the predictions are improved by the final module, the sequence-to-sequence network. In the example shown here, it corrects the words {\em lie,} {\em if,} and {\em or}.
    }
    \label{fig:overall-architecture}
    \end{figure*}

  \vspace*{-0.3cm}
\section{Related Work}
Text in images of natural scenes appears in various fonts and colors; it may be illuminated unevenly and appear in complex backgrounds.  These are all challenges that must be addressed.
We here briefly discuss related works on text recognition, sequence-to-sequence methods, and spelling correction methods.


{\bf Text recognition} is an active research topic in computer vision and document image analysis. A predominant state-of-the-art methodology is to find image regions with text and then consider one character at a time in these regions to recognize the text.  Following this methodology, He et al.~\cite{7442550}, for example, proposed a multi-layer network architecture that is unique in its convolutional input layer and processes small image patches (32x32x3) in order to recognize characters in text regions.  
The focus on characters has been considered a limitation of the methodology.  

An alternative approach is 
 try to recognize words, one at a time, once text regions have been identified. 
 Jaderberg et al.~\cite{Jaderberg2016} followed this methodology by first detecting candidate text regions with a high-recall, low-precision region-proposal network, then filtering and refining the results with a Random Forest approach to yield word regions, and finally passing these word images into a dictionary-based CNN that outputs the likelihood that the input is an image of a word in the dictionary.        
 Dictionary-based CNN classifiers have been criticized for causing severe overfitting problems if the dimension of the output one-hot vector is too large compared to the information provided by an insufficiently-sized training dataset.

{\bf  A sequence-to-sequence learning model} for text recognition was proposed by Shi et al.~\cite{ShiBaYa15}. In this framework, convolutional features are extracted at encoder stage, and then Bidirectional LSTM is applied to decode these features, then CTC is used to form the final text. Later they also developed an attention-based STN for rectifying text distortion, which is useful to recognize curved scene text (Shi et al. 2016~\cite{Shi_2016_CVPR}). Bai et al.~\cite{Bai_2018_CVPR} proposed EP text recognition model focusing on the missing, superfluous and unrecognized characters and alleviating misalignment problem for text recognition. However, these models all take texts in images as isolated individuals. Even though lexicon-based methodology is implemented, tree-based search algorithm can decrease the speed of framework.

Aside from text recognition, the sequence-to-sequence learning model is also widely used in the field of Natural Language Processing (NLP)~\cite{DBLP:journals/corr/BahdanauCB14,DBLP:journals/corr/ChoMGBSB14,
DBLP:journals/corr/JeanCMB14, DBLP:journals/corr/abs-1805-03989, DBLP:journals/corr/abs-1708-06426,NIPS2014_5346,
DBLP:journals/corr/VinyalsKKPSH14, DBLP:journals/corr/VinyalsL15}. Some popular implementations include dialogue systems~\cite{DBLP:journals/corr/VinyalsL15}, neural machine translation~\cite{DBLP:journals/corr/ChoMGBSB14, DBLP:journals/corr/JeanCMB14}, and automatic text summarization~\cite{DBLP:journals/corr/abs-1805-03989}.


{\bf Spelling error correction} is a longstanding NLP problem that has been addressed in various ways, for example, with a k-gram-based approach ~\cite{sundby2009spelling}, a text-level maximum likelihood approach~\cite{farra2014generalized}, and a sequence-to-sequence-based approach~\cite{DBLP:journals/corr/abs-1709-06429}. 

Based on the prior work on text recognition and spelling error correction, we implemented a text recognition model that provides prediction baseline results, then the proposed TGA algorithm first groups and then arranges isolated text regions within the image together. Finally, a sequence-to-sequence model is proposed to improve the text recognition accuracy based on the semantic information extracted from these text regions.

\section{Methodology}
\subsection{Overview}
The overview of our model is illustrated in Fig.~\ref{fig:overall-architecture}. STR consists of two branches and four modules, connected in series, to recognize text in images by understanding context. The four modules are a perspective transformation module, a text recognition module, a text grouping and arranging module, and a text spelling correction module.

\subsection{Perspective Transformation}
Text recognition is sensitive to text skew and distortion in images. A perspective transformation process is introduced to convert the quadrilateral text proposals into axis-aligned text regions, which makes it easier to interpret by our subsequent text recognition modules. Each text image region firstly was computed by cropping from original images and relying on the coordinates of the ground-truth bounding boxes. Second, a ``Perspective Transformation Algorithm'' is introduced to rectify text regions, which makes good foundation for text recognition.


\begin{table}
\begin{center}
\begin{tabular}{|p{30mm}|p{40mm}|}
\hline
Layer Type                & Layer Configuation                        \\ \hline\hline
conv\_bn\_relu      & k: {[}3, 3{]}; s: 1; p: 1; o: 64     \\ \hline
max-pool            & w: {[}2, 2{]}; s: 2;                 \\ \hline
conv\_bn\_relu      & k: {[}3, 3{]}; s: 1; p: 1; o: 128    \\ \hline
max-pool            & w: {[}2, 2{]}; s: 2;                 \\ \hline
conv\_bn\_relu      & k: {[}3, 3{]}; s: 1; p: 1; o: 256    \\ \hline
batch-normalization & -                                    \\ \hline
conv\_bn\_relu      & k: {[}3, 3{]}; s: 1; p: 1; o: 256     \\ \hline
max-pool            & k: {[}2, 2{]}; s: {[}2, 1{]}; p: {[}0, 1{]};   \\ \hline
conv\_bn\_relu      & k: {[}3, 3{]}; s: 1; p: 1; o: 512      \\ \hline
batch-normalization & -                                      \\ \hline
conv\_bn\_relu      & k: {[}3, 3{]}; s: 1; p: 1; o: 512                  \\ \hline
max-pool            & k: {[}2, 2{]}; s: {[}2, 1{]}; p: {[}0, 1{]};       \\ \hline
conv\_bn\_relu      & k: {[}3, 3{]}; s: 1; p: 1; o: 512                  \\ \hline
batch-normalization & -                                      \\ \hline
map-to-sequence     & -                                      \\ \hline
bi-directional LSTM & hidden units: 256                      \\ \hline
bi-directional LSTM & hidden units: 256                      \\ \hline
\end{tabular}
\end{center}
\vspace*{-0.3cm}
\caption{The configuration of the text recognition network model. The abbreviations 'k,' 's,' 'p,' 'o,' and 'w' refer to kernel size, stride size, padding size, number of output channels, and the max-pooling window size,  respectively.}
\label{table:backbone}
\end{table}

\subsection{Text Recognition}
The Text Recognition branch is to predict text's label of each letter. Inspired by CRNN \cite{ShiBaYa15}, CNN and Bidirectional LSTM are adopted to get text recognition results from text images. For the convolutional part, VGG-based convolutional layers~\cite{SimonyanZi14} automatically extract a feature sequence from each input image. For the recurrent part, A bidirectional LSTM model is built for making prediction for each frame of the feature sequence, getting predicted results for text recognition. The backbone of text recognition branch is shown in Table~\ref{table:backbone}.

At the end of text recognition branch, connectionist temporal classification (CTC) \cite{Graves:2006:CTC:1143844.1143891} is used to estimate a sequence probability, translating perfume results into a label sequence. The formulation of the conditional probability is briefly described as follows: With the input sequence \textbf{x} = $x_1$,...,$x_M$, where $M$ is the sequence length, the distribution can be given as $\mathbf{x_{m}} \in R^{|C|}$, where $C$ include all character labels. Given the ground truth sequence $\mathbf{y} = {y_1,...,y_T}$, the conditional probability of the label $\mathbf{y^*}$ is the sum of probabilities of all paths $\pi$, i.e.,
\begin{equation}
p(\mathbf{y^*}|\mathbf{x}) =\sum_{\pi \in 
\beta^{-1}(\mathbf{y}^*)} p(\pi|\textbf{x}),
\end{equation}
where 
$\beta$ defines the conditional probability given labelling $\mathbf{y}^*$ as the sum of the probabilities of all the many-to-one mapping. The training process attempts to maximize the log-likelihood of the summation in Eq.~(1) over the whole training set. Following the work by Graves et al.~\cite{Graves:2006:CTC:1143844.1143891}, we define the text recognition loss function to be
\begin{equation}
{\rm Loss}_{\rm recog} = -\frac{1}{N}\sum_{n=1}^{N} \log \left (  p(\mathbf{y}_{n}^{*}|\mathbf{x})\right ),
\end{equation}
where $N$ is the number of text regions in an input text image and $\mathbf{y}_{n}^{*}$ is the most probable label for text recognition. 

\begin{figure*}
\begin{center}
 \includegraphics[width=0.9\linewidth]{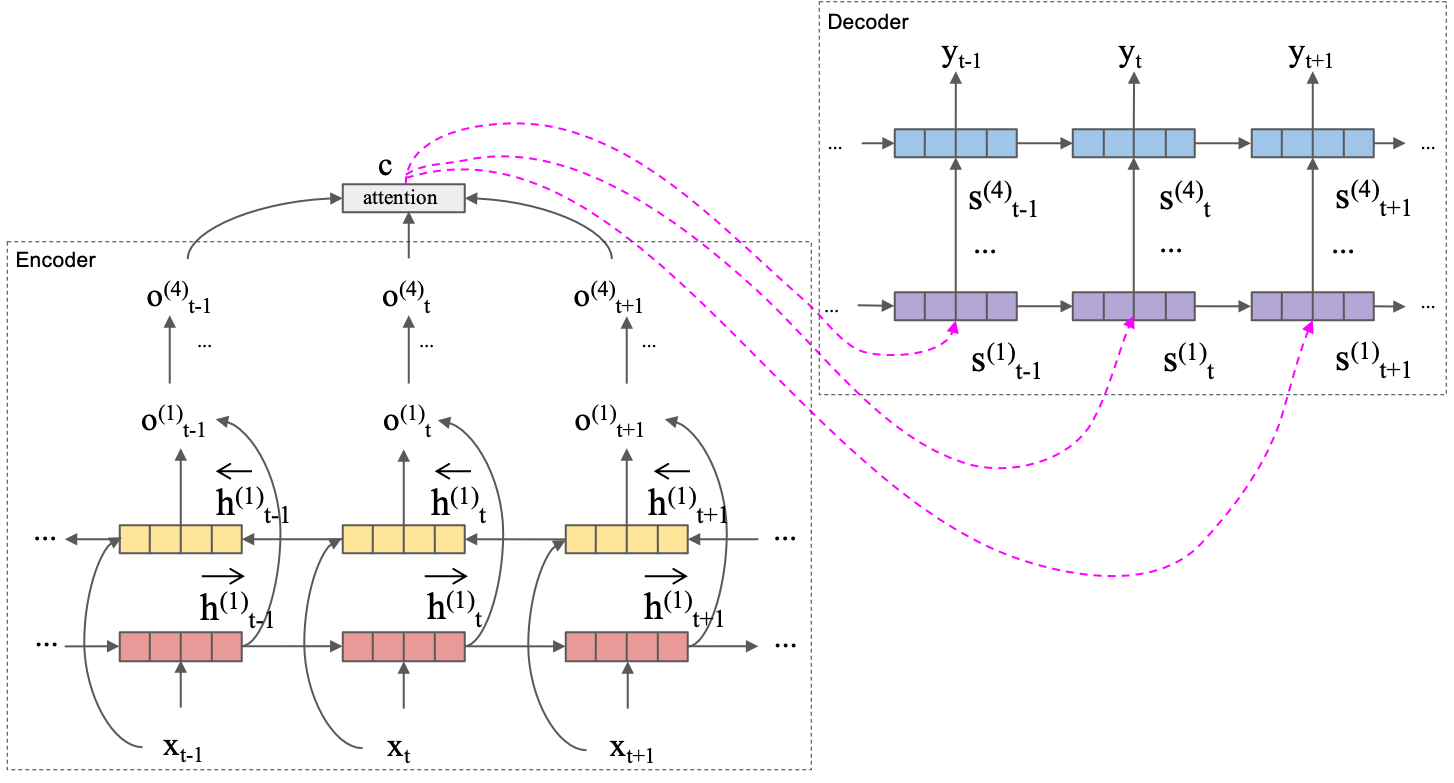}
\end{center}
\vspace*{-0.8cm}
  \caption{The structure of the proposed sequence-to-sequence spelling correction neural network. The encoder is composed of four bidirectional LSTM layers. The decoder is composed of four LSTM\cite{hochreiter1997long} layers. Global attention~\cite{DBLP:journals/corr/LuongPM15} is used to reduce the information loss from the encoder.}
\label{fig:structure-seman-spell}
\end{figure*}

\subsection{Text Grouping and Arranging}

In order to retrieve the semantic information of a text in an image, it is essential to group and arrange isolated text or word regions into paragraphs or sentences in the correct logical order. To accomplish this, we propose the Text Grouping and Arranging (TGA) algorithm. The TGA algorithm takes geometric information of each text region from the Text Detection module, and outputs the text of logically grouped and arranged sentences or paragraphs. It uses two processes, text grouping and text arranging, which are explained in detail below. Sample results of the TGA algorithm are shown in Fig.~\ref{fig:result_tpid}.



\subsubsection{Text Grouping}
The core idea of the text grouping process is inspired by the flood fill algorithm. The text grouping process starts from a randomly selected text bounding box and groups neighboring bounding boxes together. We assume that these neighboring text regions belong to one sentence or paragraph. The details of the text grouping algorithm are summarized in Algorithm~\ref{Text Grouping}. In the pseudocode, $U$ denotes the unlabeled text bounding boxes, and $L$ represents the labeled text bounding boxes. Due to the text grouping process, we can successfully group text regions from images by relying on their geometric information.

\begin{algorithm}
  \begin{algorithmic}[1]
  \Require Unlabeled Text Bounding Boxes: $U$
  \Ensure Labeled Text Bounding Boxes: $L$ 
    \Function{Grouping}{$U$}
    \State $L\gets \oldemptyset; L'\gets \oldemptyset ;label\gets 0$
    \State $(b_i, label = label + 1)\gets$ \textbf{Dequeue}$(U)$ 
    \While{$L'\not=\oldemptyset$}
      \State $(seed, label)\gets$ \textbf{Dequeue}$(L')$ 
      \State $U'\gets U; U\gets \oldemptyset$
      \While{$U'\not=\oldemptyset$}
        \State $b_i \gets$ \textbf{Dequeue}$(U')$
        \If {\textbf{SameGroup}($seed, b_i$)}
        \State \textbf{Enqueue}$(L',(b_i,label))$
        \State $L\gets L \cup \{(b_i, label) \}$
        \Else {} \textbf{Enqueue}$(U,b_i)$
        \EndIf
      \EndWhile
      \If {$U\not=\oldemptyset$ and $L' =\oldemptyset$}
      \State $(b_i, label = label + 1)\gets$ \textbf{Dequeue}$(U)$ 
      \State \textbf{Enqueue}$(L',(b_i,label))$
      \State $L\gets L \cup \{(b_i, label) \}$
      \EndIf
    \EndWhile
    \State \textbf{return} $L$
    \EndFunction
  \end{algorithmic}
  \caption{Algorithm for Text Grouping }
  \label{Text Grouping}
\end{algorithm}

\subsubsection{Text Arranging}
The next step of the TGA algorithm is to arrange grouped texts in the correct logical order with the help of the location of text bounding boxes in the images.

The text arranging process is inspired by the Linked List Traversing Algorithm. We assume that the correct text arrangement starts from the left and continues to the right, and always proceeds from the top to the bottom.  Therefore all the grouped text bounding boxes are first sorted according to their horizontal coordinates then their vertical coordinates. The details of the text arranging process are summarized in Algorithm~\ref{Text Arranging}. In the pseudocode, $U$ denotes unarranged text bounding boxes. $A$ represents arranged text bounding boxes. The text arranging process enables us to arrange text regions in the correct logic order. 

\begin{algorithm}
  \begin{algorithmic}[1]
  \Require Unarranged Text Bounding Boxes: $U$
  \Ensure Arranged Text Bounding Boxes: $A$ 
    \Function{Arranging}{$U$}
    \State $L\gets \oldemptyset;$
    \For{$u_i \in U$}
      \State $l\gets \oldemptyset;$
      \State $p_{\rm current}\gets$\textbf{ GetLocation}$(u_i)$
        \While{{\em True}}
          \State $l\gets l\cup \{$\textbf{GetValue}$(p_{\rm current})\}$
          \State $p_{\rm next}\gets $ \textbf{FindNexttext}$(p_{\rm current})$
          \State \Comment{Find pointer to next region on same line}
          \If{$p_{next} = $\textbf{ null}}
            \State \textbf{Break}
          \EndIf
          \State $p_{\rm current}\gets p_{\rm next}$
        \EndWhile
      \State $L\gets L\cup \{l\}$
    \EndFor
    \State $A\gets \oldemptyset;$
    \State $A\gets $\textbf{SortByVerticalLocation}$($\textbf{DeleteSubLine}$(L)$$)$\Comment{Only keep list containing the most text regions in one line and sort relying on vertical coordinates in one line}
    \State \textbf{return }$A$
    \EndFunction
  \end{algorithmic}
  \caption{Algorithm for Text Arranging}
  \label{Text Arranging}
\end{algorithm}
\subsection{Spelling Correction}

A sequence-to-sequence model is designed to correct spelling error by relying on semantic information in a specific domain. The backbone of the proposed model is based on a sequence-to-sequence network~\cite{NIPS2014_5346} that has produced high-accuracy results for many NLP problems, including analyzing dialogues~\cite{AsriHeSu16}, neural machine translation\cite{DBLP:journals/corr/WuSCLNMKCGMKSJL16}, and automatic text summarization~\cite{NallapatiXiZh16,NallapatiZhSaGuXi16}. These models learn to generate a variable-length information sequence or ``flow of tokens'' (e.g., an English sentence) from a variable-length input sequence or ``input flow'' (e.g., the corresponding Chinese sentence). LSTMs are used in the sequence-to-sequence framework to handle the exploding and vanishing gradient problems and make the best of the information about the immediate image background of the characters in an input sequence and their fonts. 

The encoder of the sequence-to-sequence framework compresses the entire sequence of information into a fixed-length context vector, which cannot fully represent the information of the entire sequence, especially when long input sequences are given. Fortunately, this problem has been solved by the attention model~\cite{DBLP:journals/corr/BahdanauCB14}. It generates a ``focus range''  to indicate which parts of the input sequence should be focused on, and then produces the next output sequences based on the region of focus. The attention-based sequence-to-sequence model enables focus on specific parts of the input automatically to help generate a more accurate output result.

Before training our model, we need to transform the input sequences into a form that our sequence-to-sequence model can understand. Some examples are shown in Table~\ref{table:Example-input-data}. We treat each character as one word, then split each text using grave accent (`) and split each character using space. The beginning of the sequence is marked as $\langle GO \rangle$, and the end of the sequence is marked as $\langle END \rangle$.

\begin{table}
\begin{center}
\begin{tabular}{|l|c|}
\hline
Original sequence & Input sequence \\
\hline\hline
sitting room & s i t t i n g ` r o o m \\
black or yellow-red & b l a c k ` o r ` y e l l o w - r e d \\
respect for all & r e s p e c t ` f o r ` a l l \\
all lifes matter & a l l ` l i f e s ` m a t t e r \\
\hline
\end{tabular}
\end{center}
\vspace*{-0.3cm}
\caption{Examples of input sequence pre-processing.}
\label{table:Example-input-data}
\end{table}

The structure of our sequence-to-sequence spelling-correction model is shown in Fig.~\ref{fig:structure-seman-spell}. In this framework, the input sequence $\mathbf{x}$ is transferred into a context vector sequence $c$ by a bidirectional LSTM-based encoder. The encoder process \cite{DBLP:journals/corr/abs-1810-00660} can be defined as
\begin{equation}
\begin{split}
    \overrightarrow{h_t} = f(x_t, \overrightarrow{h_{t-1}})\\
    \overleftarrow{h_t} = f(x_t, \overleftarrow{h_{t-1}}),
\end{split}
\end{equation}
where $\{\overrightarrow{h_1},...,\overrightarrow{h_t},\overleftarrow{h_1},...,\overleftarrow{h_t}\}\in \mathbb{R}^{2t}$ are the encoder hidden states at time $t$.

The decoder is trained to predict the next character~$y_t$, given the context vector sequence $c$ and all the previously predicted characters $y_1, y_2,...,y_{t-1}$. The probability for the output sequence $\mathbf{y}$ can be defined as
\begin{equation}
    p(\mathbf{y}) = \prod_{t = 1}^{T} p(y_t|y_1,...,y_{t-1}, c),
    \label{Equ decoder}
\end{equation}
where the conditional probability can be defined as
\begin{equation}
    p(y_t | y_1,..., y_{t-1}, \mathbf{x}) = g(y_{t-1}, s_t, c_t),
\end{equation}
with 
$s_t = f(s_{t-1}, y_{t-1}, c_t)$ 
denoting an decoder hidden state at time~$t$.

In the attention-based sequence-to-sequence model~\cite{ DBLP:journals/corr/BahdanauCB14, DBLP:journals/corr/LuongPM15}, the mapping from each context vector $c_i$ to the encoder hidden state $\{\overrightarrow{h_1},...,\overrightarrow{h_t},\overleftarrow{h_1},...,\overleftarrow{h_t}\}$ is computed as
\begin{equation}
    c_i = \sum_{j = 1}^{t}\alpha_{ij}\overleftarrow{h_j} + \sum_{j = 1}^{t}\alpha_{ij}\overrightarrow{h_j}.
\end{equation}
The weight $\alpha_{ij}$ for each $h_j$ can be computed by 
\begin{equation}
    \alpha_{ij} = \frac{\exp(e_{ij})}{\sum_{k=1}^{T_x}\exp(e_{ik})},
\end{equation}
where 
$e_{ij} = a(s_{i-1}, h_j)$ 
scores matching degree between input character~$j$ and output character~$i$.

During the training process, the training loss is computed on the output of the decoder at the character level. At the each time step $t$, the implemented loss function is the cross-entropy loss per time step that is relevant to the previous time step~\cite{DBLP:journals/corr/abs-1709-06429}, which is computed as 
\begin{equation}
    {\rm Loss}(x, y) = -\sum_{t = 1}^{T}\log(P(y_t|x, y_{t-1}, y_{t-2},...,y_1)).
\end{equation}

\section{Experiments}

We evaluate the proposed STR method on two benchmark datasets and compare our results to a baseline method.  We first explain why it is not meaningful to compare our model to various existing text-spotting methods.  A comparison to a single baseline method, which uses state-of-the-art isolated-word recognition techniques, is sufficient.

\subsection{Baseline Method}

The performance of a new method is ideally tested on one or more existing benchmark dataset(s), which allows comparison to the performance of state-of-the-art methods on the same data.  The benchmark datasets used to test text recognition methods are ICDAR 2013~\cite{karatzas2013icdar}, which contains 1,015 images of cropped words, III5k-words~\cite{mishra2012scene}, which contains 3,000 test images collected from the Web, and/or Street View Text~\cite{wang2011end}, collected from Google Street View (and thus mostly contains sparse text regions).   It is important to note that the text in these image datasets are typically isolated words, which are not relevant to each other semantically. So the text in the image cannot express unified semantic information. STR relies on potential semantic information to improve the prediction of our text recognition module, which uses state-of-the-art isolated-word recognition.  In order to test the benefit of using semantic information in STR, we therefore work with newly created datasets, which we describe in detail next. 


\subsection{Benchmark Datasets}

We introduce the \textbf{Interior Design Dataset} (IDD).  It consists of 7,708 images of scanned product catalogues for interior design and decoration.  
We select 4,708 of them as training images, 1,500 as validation images, and the remaining 1,500 for testing. The document images used for training contain more than 600,000 image regions with text and the number of image regions with text used for testing is 251,074. The ground truth labels are the bounding-box coordinates of each image region that contains a word and a textual representation of the word itself.

We created a subset of the UCLA Protest Image Dataset~\cite{won2017protest}, which is a collection of social media images that can be used to detect protest activities in street scenes and evaluate the potential of violence in these protest activities.  The orginal dataset consists of 40,764 images, among which there are 11,659 images showing a protest.
Among the protest images, we identified 816 images that contain mostly hand-made signs and text that is hand written and select 656 of them as training images and 160 as testing images. We created ground-truth bounding boxes and textual representations of all the words on every sign. We refer to the protest-sign image collection as the {\bf Text-containing Protest Image Dataset} (TPID). The total number of words for testing is 2,293.

\subsection{Training Strategy}

To analyze the performance of our STR method on the two datasets IDD and TPID, we train the text recognition and sequence-to-sequence spelling correction models specifically on each dataset, yielding a IDD-specific STR system that can recognize text in scanned pages of interior design catalogues, and a TPID-specific STR system that can recognize text (typically hand-drawn) on protest signs in natural scenes.   We employ different training strategies for the two experiments: 

Our text recognition model is trained on MJSynth~\cite{Jaderberg14d,Jaderberg14c} and SynthText~\cite{Gupta16} datasets for TPID STR system. For the IDD STR system, we fine-tune this pre-trained model using 4,708 training images from IDD. These training images contains more than 600,000 text image regions. We adopt RMSprop as the optimizer for this IDD text-recognition model and set the learning rate to 0.0001.

The spelling-correction branch is trained by using the OpenNMT-py framework~\cite{opennmt} with batch size equal to~64. the training dataset consists of grouped and arranged sentences from IDD obtained with our TGA algorithm. Our encoder is composed of a 4-layer bidirectional LSTM and our decoder is composed of 4-layer LSTM with a dropout rate set to 0.3.  Encoder and decoder are connected with the Luong Attention framework~\cite{DBLP:journals/corr/LuongPM15}. The optimization method used by our IDD spelling-correction model is stochastic gradient descent. Its learning rate is set to 1.0, starting decay from 50,000 steps and cutting in half every 10,000 steps. For the TPID STR system, due to limited number of training images, we first get a pre-trained spelling-correction model using titles related to protest from the Thompson dataset~\cite{thompson_2017} with the same parameters and process as IID STR systems. We then manually induce noise for grouped and arranged sentences of training images to fine-tune it. 

The STR system is implemented using Pytorch on a single NVNDIA GTX 1080Ti graphic card with 11~GB memory.



\subsection{Experimental Results and Discussion}

  The proposed STR method recognizes 90\% (=226,067/251,074) of the words in IDD and 71\% (=1,630/2,293) of the words in TPID correctly.  Our results reveal that it outperforms the text recognition baseline for both datasets. In particular, on IDD, STR beats the baseline by   4.68 percent points and, on TPID, STR beats the baseline by 5.46 percent points
(Table~\ref{table:overall-result}).

\begin{table}
\begin{center}
\begin{tabular}{|l|c|c|}
\hline
 & Design (IDD) & Protest (TPID) \\
\hline
Baseline & 85.36 \% & 65.63 \% \\
Proposed STR  & \textbf{90.04 \%} & \textbf{71.09 \%} \\
\hline
\end{tabular}
\end{center}
\vspace*{-0.3cm}
\caption{Results of our STR method applied to two datasets and compared to the baseline method, which uses state-of-the-art text recognition techniques.}
\label{table:overall-result}
\end{table}

\begin{figure}
\begin{center}
    \includegraphics[width=\columnwidth]{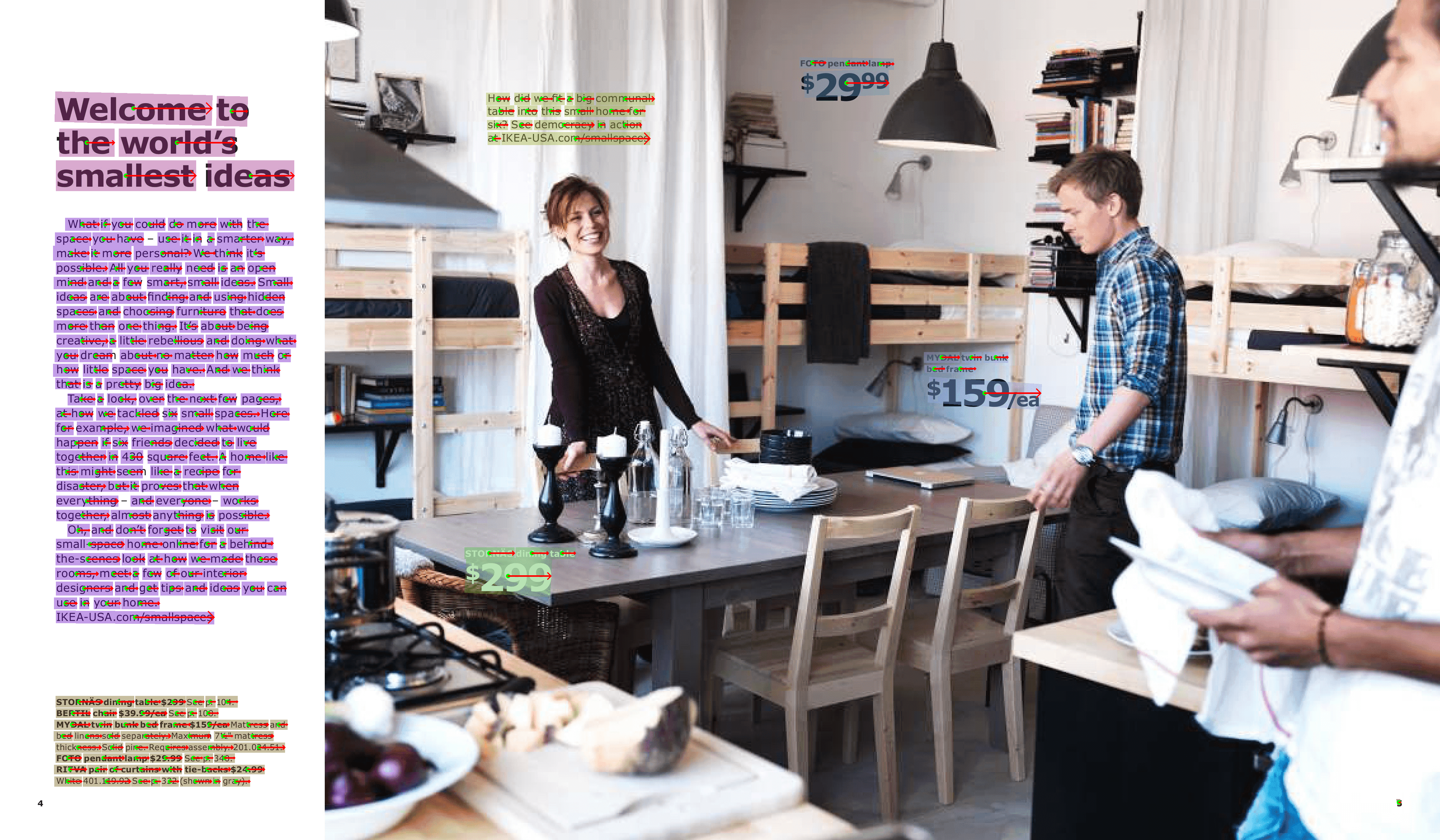}
\end{center}
\caption{Results of TGA on a document image from the interior design dataset.  
Texts that TGA grouped into a single semantic context are visualized with the same background color (seven regions in this example). The center of each recognized word is shown with a green dot. A red arrow indicates  connection of the words within a text row and across text rows.  
}
\label{fig:result_idd}
\end{figure}

\begin{figure}
    \begin{subfigure}{0.4\linewidth}
        \includegraphics[height=40mm, width=40mm]{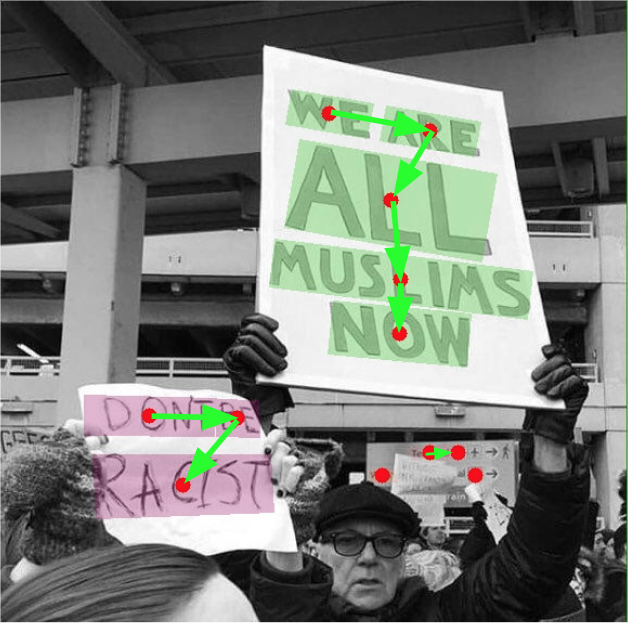}
        \caption{}
        \label{5a}
    \end{subfigure}
    \hspace{2em}
    \begin{subfigure}{0.4\linewidth}
        \includegraphics[height=40mm, width=40mm]{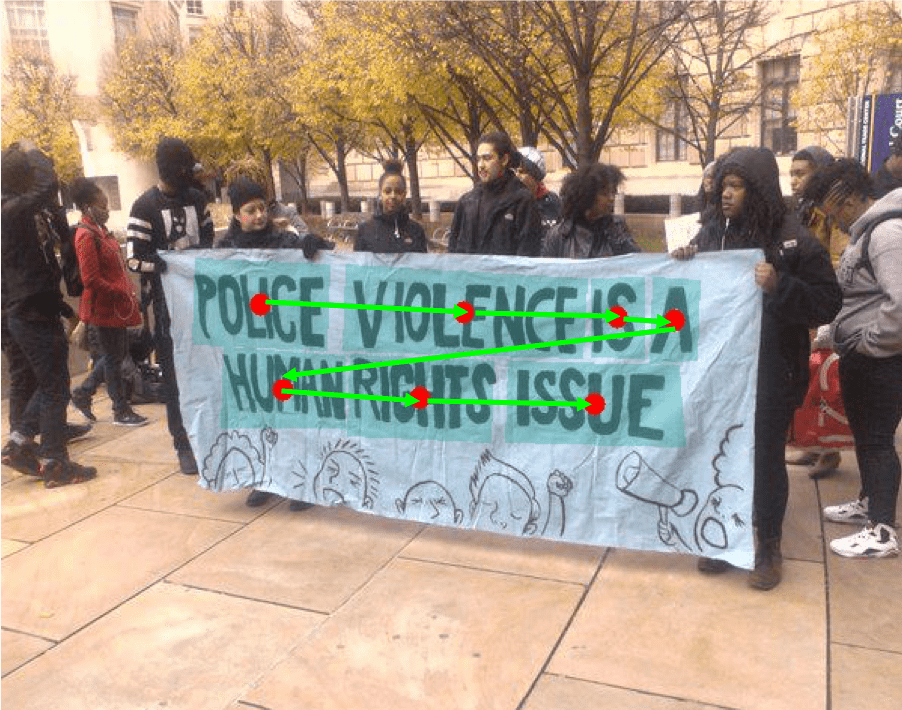}
        \caption{}
        \label{5b}
    \end{subfigure}%
    
    \begin{subfigure}{0.4\linewidth}
        \includegraphics[height=40mm, width=40mm]{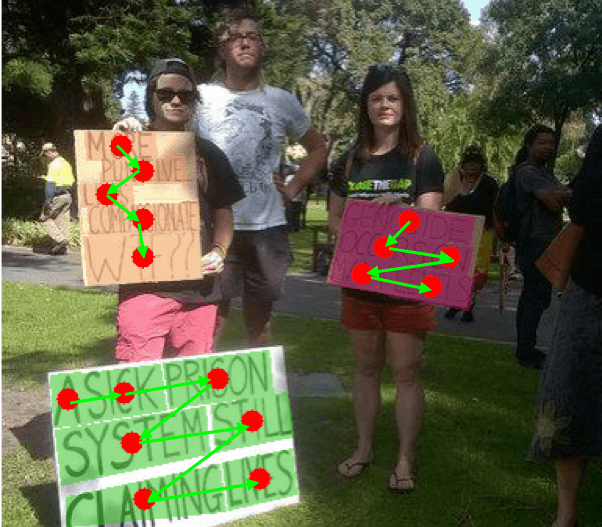}
        \caption{}
        \label{5c}
    \end{subfigure}
    \hspace{2em}
    \begin{subfigure}{0.4\linewidth}
        \includegraphics[height=40mm, width=40mm]{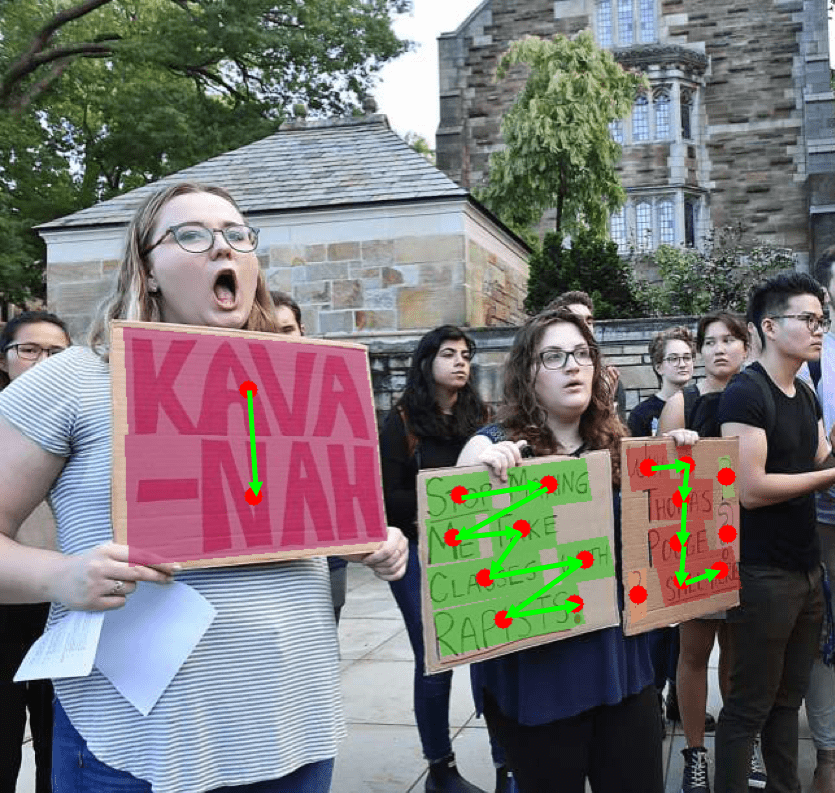}
        \caption{}
        \label{5d}
    \end{subfigure}%
    
    \begin{subfigure}{0.4\linewidth}
        \includegraphics[height=40mm, width=40mm]{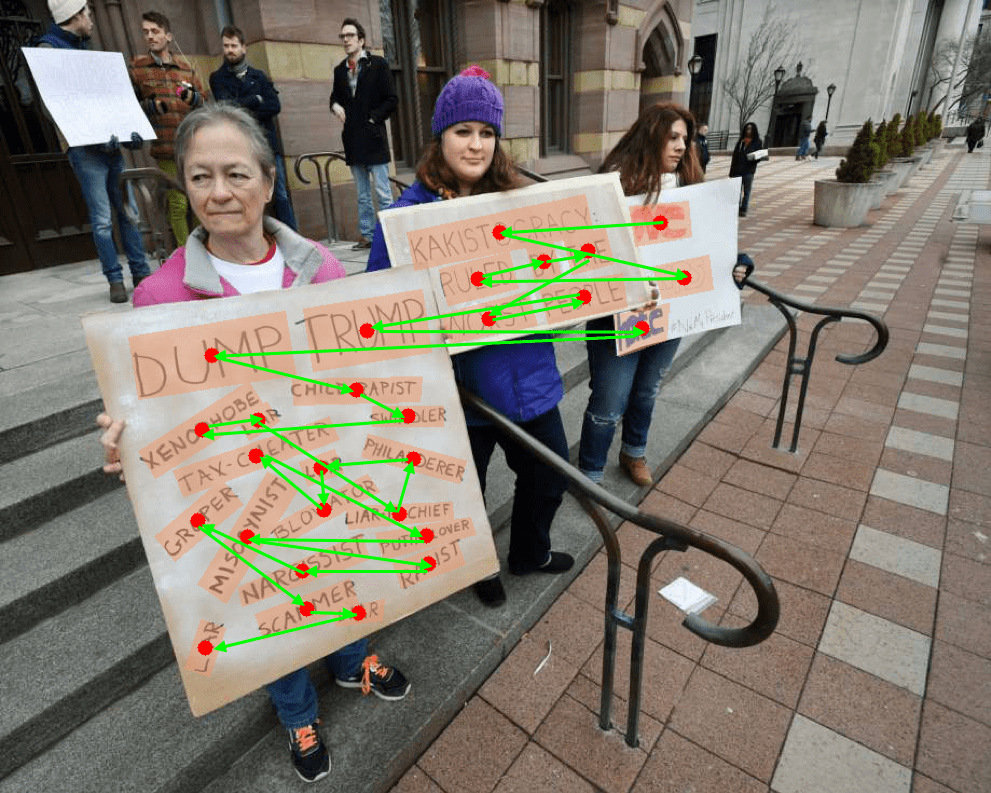}
        \caption{}
        \label{5e}
    \end{subfigure}
    \hspace{2em}
    \begin{subfigure}{0.4\linewidth}
        \includegraphics[height=40mm, width=40mm]{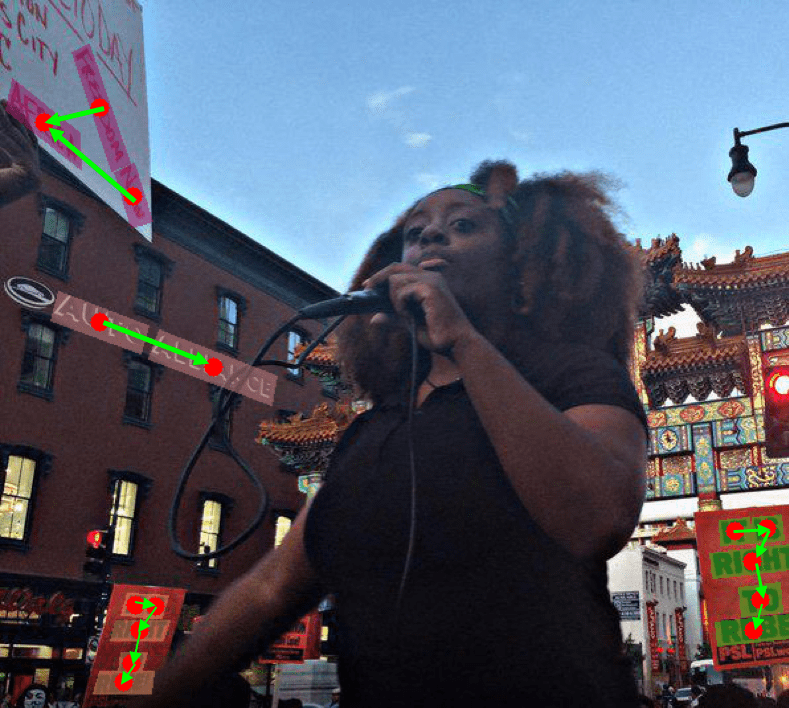}
        \caption{}
        \label{5f}
    \end{subfigure}%
    \caption{Results of TGA on TPID images. 
    Perfect results of TGA are shown in (a)--(d), failure cases are shown in (e) and (f). The center of each recognized word is shown as a red dot. A green arrow indicates the semantic connection of the words within a text region. It is noteworthy that, in (d), the words on the two neighboring signs, held up by the same woman, are correctly recognized to belong to different contexts. This does not apply to the three signs in (e), which contain a collection of disjoint, rotated words, and for which the semantic context is difficult to establish. Since only a part of the sign in the top left corner of the image in (f) is shown, TGA failed to order the words on the sign correctly. 
    %
    }
\label{fig:result_tpid}
\end{figure}

The high accuracy rate of STR on the document images is due to the fact that the task is relatively easy. The words are all machine generated, horizontally aligned, and the characters within a word have fixed fonts.  This reduces the difficulty of text recognition, grouping and arranging. The task of interpreting text in the protest images is significantly more difficult. Due to the diversity of handwritten text layouts, text grouping in TPID is not easy. 
At the same time, the variability of the aspect ratio and font type of the handwritten text also makes it more challenging to arrange and recognize the text on the signs in the TPID.

The TGA algorithm can play an important role in text recognition because it groups and arranges words in images. The performance of TGA on the document image data IDD was excellent (Fig.~\ref{fig:result_idd}).  

The performance of TGA on the protest image data TPID was reliable in images with single, clearly-written signs (
(Fig.~\ref{fig:result_tpid} (a)--(c)) and not as strong if several difficult-to-read text regions are present  (Fig.~\ref{fig:result_tpid} (e)--(f)). Words belonging to different signs that are very close to each other rises the difficulty of grouping and causes the text regions that do not belong to each other be grouped together. This phenomenon is rare in IDD. An additional hurdle is that, due to the diversity of the size and font of the handwritten text on protest signs, the bounding boxes of the words in TPID as not as clearly aligned horizontally as in the IDD.  This causes a higher error rate in text arranging.

\begin{figure}
    \begin{subfigure}{0.4\linewidth}
        \includegraphics[height=40mm, width=40mm]{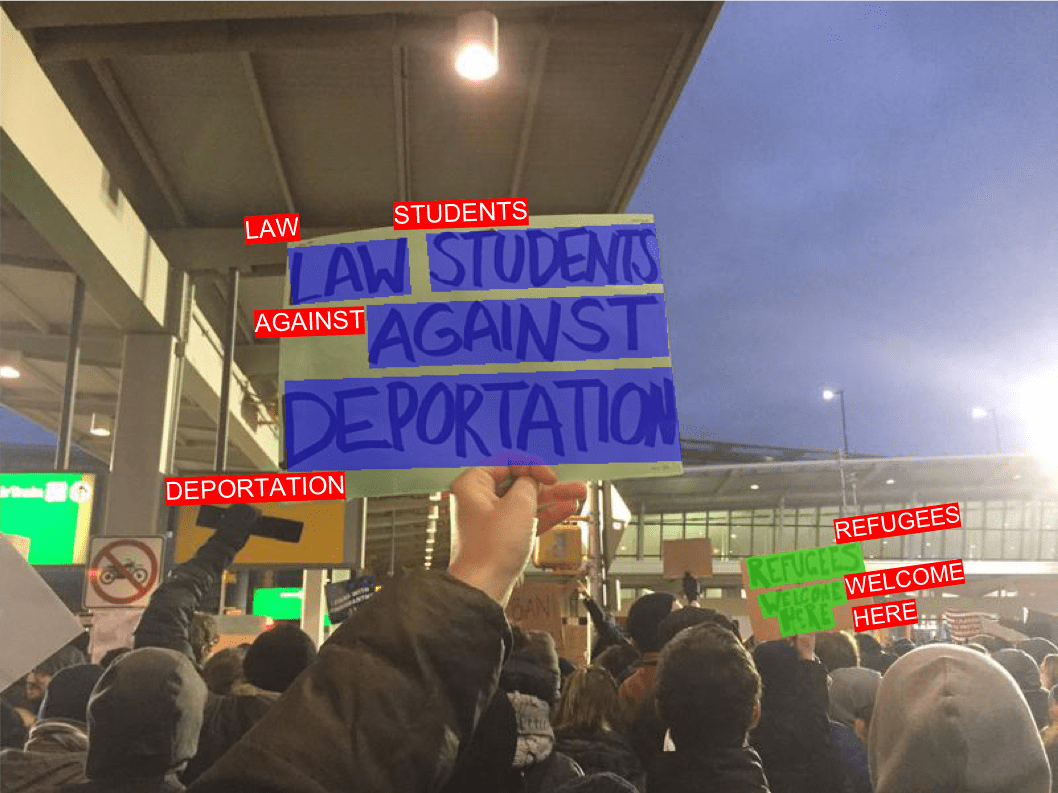}
        \caption{}
        \label{6a}
    \end{subfigure}
    \hspace{2em}
    \begin{subfigure}{0.4\linewidth}
        \includegraphics[height=40mm, width=45mm]{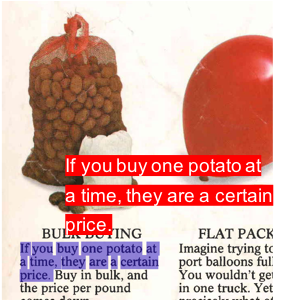}
        \caption{}
        \label{6b}
    \end{subfigure}%
    \caption{Sample results of STR.  In the TPID image in (a), the baseline text recognition method outputs 'REFUGEES WELCONE HERE,' while STR  predicts the correct 'REFUGEES WELCOME HERE.' In the IID image in (b), the baseline method omits two occurrences of the word 'a,' but STR correctly predicts these two words based on the context of the sentence.}
    \label{fig:STR-results}
\end{figure}

Compared to the text recognition baseline, STR performs more accurately because it can use the overall context information from the text regions in an image to check and correct predicted results from the single-word text-recognition baseline (Figs.~\ref{fig:task} and~\ref{fig:STR-results}). At the same time, we found that STR can utilize morphological information of one text to correct the prediction from the text recognition baseline. STR combines multi-level information, instead of image information, and thus can improve the accuracy of the text recognition baseline. Note that combination of multi-modal information to obtain improved prediction results is of great significance to the research work in the field of computer vision.


\section{Conclusions}

In this work, we proposed a new deep learning model called STR. This model can efficiently understand the context between regions of text or between words in images. By grouping and arranging text and relying on semantic information from images, our model is able to efficiently improve prediction results from text recognition and extract sentences or paragraphs from images instead of isolated text regions or words like state-of-the-art frameworks do. The experiments demonstrate that our model achieves superior or highly competitive performance and suggest generality in that our model can handle different semantic contexts in images.

We achieved this performance based on two insights -- we realized the value of grouping text that belongs together and the value of correcting word choices and spelling with contextual information.  We proposed the TGA algorithm to group and arrange text in the same paragraph together, and we succeeded in applying spelling correction with contextual information. 

Our datasets, their ground-truth labels, as well as our code will be made publicly available with publication of this work.  We hope to inspire follow-up work that tries to improve upon the performance of text recognition in the challenging protest image dataset, and also 
applies our ideas to text recognition in other domains.

\section*{Acknowledgements}

This work has been partially supported by the National Science Foundation, grant 1838193 (to M.B.).

{\small
\bibliographystyle{ieee}
\bibliography{text-recognition}

\begin{thebibliography}{10}\itemsep=-1pt

\bibitem{DBLP:journals/corr/abs-1810-00660}
S.~Ahmadi.
\newblock Attention-based encoder-decoder networks for spelling and grammatical
  error correction.
\newblock {\em CoRR}, abs/1810.00660, 2018.

\bibitem{AsriHeSu16}
L.~E. Asri, J.~He, and K.~Suleman.
\newblock A sequence-to-sequence model for user simulation in spoken dialogue
  systems.
\newblock In {\em Interspeech 2016, September 8–-12, 2016, San Francisco,
  USA}, pages 1151--1155. ISCA, 2016.
\newblock http://arxiv.org/abs/1607.00070.

\bibitem{DBLP:journals/corr/BahdanauCB14}
D.~Bahdanau, K.~Cho, and Y.~Bengio.
\newblock Neural machine translation by jointly learning to align and
  translate.
\newblock {\em CoRR}, abs/1409.0473, 2014.

\bibitem{Bai_2018_CVPR}
F.~Bai, Z.~Cheng, Y.~Niu, S.~Pu, and S.~Zhou.
\newblock Edit probability for scene text recognition.
\newblock In {\em The IEEE Conference on Computer Vision and Pattern
  Recognition (CVPR)}, June 2018.

\bibitem{DBLP:journals/corr/ChoMGBSB14}
K.~Cho, B.~van Merrienboer, {\c{C}}.~G{\"{u}}l{\c{c}}ehre, F.~Bougares,
  H.~Schwenk, and Y.~Bengio.
\newblock Learning phrase representations using {RNN} encoder-decoder for
  statistical machine translation.
\newblock {\em CoRR}, abs/1406.1078, 2014.

\bibitem{farra2014generalized}
N.~Farra, N.~Tomeh, A.~Rozovskaya, and N.~Habash.
\newblock Generalized character-level spelling error correction.
\newblock In {\em Proceedings of the 52nd Annual Meeting of the Association for
  Computational Linguistics (Volume 2: Short Papers)}, volume~2, pages
  161--167, 2014.

\bibitem{DBLP:journals/corr/abs-1709-06429}
S.~Ghosh and P.~O. Kristensson.
\newblock Neural networks for text correction and completion in keyboard
  decoding.
\newblock {\em CoRR}, abs/1709.06429, 2017.

\bibitem{Graves:2006:CTC:1143844.1143891}
A.~Graves, S.~Fern\'{a}ndez, F.~Gomez, and J.~Schmidhuber.
\newblock Connectionist temporal classification: Labelling unsegmented sequence
  data with recurrent neural networks.
\newblock In {\em Proceedings of the 23rd International Conference on Machine
  Learning}, ICML '06, pages 369--376, New York, NY, USA, 2006. ACM.

\bibitem{Gupta16}
A.~Gupta, A.~Vedaldi, and A.~Zisserman.
\newblock Synthetic data for text localisation in natural images.
\newblock In {\em IEEE Conference on Computer Vision and Pattern Recognition},
  2016.

\bibitem{7442550}
T.~{He}, W.~{Huang}, Y.~{Qiao}, and J.~{Yao}.
\newblock Text-attentional convolutional neural network for scene text
  detection.
\newblock {\em IEEE Transactions on Image Processing}, 25(6):2529--2541, June
  2016.

\bibitem{hochreiter1997long}
S.~Hochreiter and J.~Schmidhuber.
\newblock Long short-term memory.
\newblock {\em Neural computation}, 9(8):1735--1780, 1997.

\bibitem{Jaderberg14d}
M.~Jaderberg, K.~Simonyan, A.~Vedaldi, and A.~Zisserman.
\newblock Reading text in the wild with convolutional neural networks.
\newblock {\em arXiv preprint arXiv:1412.1842}, 2014.

\bibitem{Jaderberg14c}
M.~Jaderberg, K.~Simonyan, A.~Vedaldi, and A.~Zisserman.
\newblock Synthetic data and artificial neural networks for natural scene text
  recognition.
\newblock {\em arXiv preprint arXiv:1406.2227}, 2014.

\bibitem{Jaderberg2016}
M.~Jaderberg, K.~Simonyan, A.~Vedaldi, and A.~Zisserman.
\newblock Reading text in the wild with convolutional neural networks.
\newblock {\em International Journal of Computer Vision}, 116(1):1--20, Jan
  2016.

\bibitem{DBLP:journals/corr/JeanCMB14}
S.~Jean, K.~Cho, R.~Memisevic, and Y.~Bengio.
\newblock On using very large target vocabulary for neural machine translation.
\newblock {\em CoRR}, abs/1412.2007, 2014.

\bibitem{karatzas2013icdar}
D.~Karatzas, F.~Shafait, S.~Uchida, M.~Iwamura, L.~G. i~Bigorda, S.~R. Mestre,
  J.~Mas, D.~F. Mota, J.~A. Almazan, and L.~P. De~Las~Heras.
\newblock {ICDAR} 2013 robust reading competition.
\newblock In {\em 2013 12th International Conference on Document Analysis and
  Recognition}, pages 1484--1493. IEEE, 2013.

\bibitem{opennmt}
G.~Klein, Y.~Kim, Y.~Deng, J.~Senellart, and A.~M. Rush.
\newblock Open{NMT}: Open-source toolkit for neural machine translation.
\newblock In {\em Proc. ACL}, 2017.

\bibitem{DBLP:journals/corr/abs-1805-03989}
J.~Lin, X.~Sun, S.~Ma, and Q.~Su.
\newblock Global encoding for abstractive summarization.
\newblock {\em CoRR}, abs/1805.03989, 2018.

\bibitem{DBLP:journals/corr/LuongPM15}
M.~Luong, H.~Pham, and C.~D. Manning.
\newblock Effective approaches to attention-based neural machine translation.
\newblock {\em CoRR}, abs/1508.04025, 2015.

\bibitem{mishra2012scene}
A.~Mishra, K.~Alahari, and C.~Jawahar.
\newblock Scene text recognition using higher order language priors.
\newblock In {\em BMVC-British Machine Vision Conference}. BMVA, 2012.

\bibitem{NallapatiXiZh16}
R.~Nallapati, B.~Xiang, and B.~Zhou.
\newblock Sequence-to-sequence {RNN}s for text summarization.
\newblock In {\em Workshop track, International Conference on Learning
  Representations (ICLR)}, pages 1--4, May 2016.

\bibitem{NallapatiZhSaGuXi16}
R.~Nallapati, B.~Zhou, C.~dos Santos, {\c{C}}.~G{\"{u}}l{\c{c}}ehre, and
  B.~Xiang.
\newblock Abstractive text summarization using sequence-to-sequence {RNN}s and
  beyond.
\newblock {\em CoRR}, abs/1602.06023:1--12, Aug. 2016.

\bibitem{schuster1997bidirectional}
M.~Schuster and K.~K. Paliwal.
\newblock Bidirectional recurrent neural networks.
\newblock {\em IEEE Transactions on Signal Processing}, 45(11):2673--2681,
  1997.

\bibitem{ShiBaYa15}
B.~Shi, X.~Bai, and C.~Yao.
\newblock An end-to-end trainable neural network for image-based sequence
  recognition and its application to scene text recognition.
\newblock arXiv:1507.05717, 2015.

\bibitem{Shi_2016_CVPR}
B.~Shi, X.~Wang, P.~Lyu, C.~Yao, and X.~Bai.
\newblock Robust scene text recognition with automatic rectification.
\newblock In {\em The IEEE Conference on Computer Vision and Pattern
  Recognition (CVPR)}, June 2016.

\bibitem{SimonyanZi14}
K.~Simonyan and A.~Zisserman.
\newblock Very deep convolutional networks for large-scale image recognition.
\newblock {\em CoRR}, abs/1409.1556, 2014.

\bibitem{DBLP:journals/corr/abs-1708-06426}
A.~Sriram, H.~Jun, S.~Satheesh, and A.~Coates.
\newblock Cold fusion: Training seq2seq models together with language models.
\newblock {\em CoRR}, abs/1708.06426, 2017.

\bibitem{sundby2009spelling}
D.~Sundby.
\newblock Spelling correction using n-grams.
\newblock {\em Technical notes}, 2009.

\bibitem{NIPS2014_5346}
I.~Sutskever, O.~Vinyals, and Q.~V. Le.
\newblock Sequence to sequence learning with neural networks.
\newblock In Z.~Ghahramani, M.~Welling, C.~Cortes, N.~D. Lawrence, and K.~Q.
  Weinberger, editors, {\em Advances in Neural Information Processing Systems
  27}, pages 3104--3112. Curran Associates, Inc., 2014.

\bibitem{thompson_2017}
A.~Thompson.
\newblock All the news: 143,000 articles from 15 {A}merican publications.
\newblock https://www.kaggle.com/snapcrack/all-the-news, Aug 2017.

\bibitem{DBLP:journals/corr/VinyalsKKPSH14}
O.~Vinyals, L.~Kaiser, T.~Koo, S.~Petrov, I.~Sutskever, and G.~E. Hinton.
\newblock Grammar as a foreign language.
\newblock {\em CoRR}, abs/1412.7449, 2014.

\bibitem{DBLP:journals/corr/VinyalsL15}
O.~Vinyals and Q.~V. Le.
\newblock A neural conversational model.
\newblock {\em CoRR}, abs/1506.05869, 2015.

\bibitem{wang2011end}
K.~Wang, B.~Babenko, and S.~Belongie.
\newblock End-to-end scene text recognition.
\newblock In {\em 2011 International Conference on Computer Vision}, pages
  1457--1464. IEEE, 2011.

\bibitem{won2017protest}
D.~Won, Z.~C. Steinert-Threlkeld, and J.~Joo.
\newblock Protest activity detection and perceived violence estimation from
  social media images.
\newblock In {\em Proceedings of the 25th ACM international conference on
  Multimedia}, pages 786--794. ACM, 2017.

\bibitem{DBLP:journals/corr/WuSCLNMKCGMKSJL16}
Y.~Wu, M.~Schuster, Z.~Chen, Q.~V. Le, M.~Norouzi, W.~Macherey, M.~Krikun,
  Y.~Cao, Q.~Gao, K.~Macherey, J.~Klingner, A.~Shah, M.~Johnson, X.~Liu,
  L.~Kaiser, S.~Gouws, Y.~Kato, T.~Kudo, H.~Kazawa, K.~Stevens, G.~Kurian,
  N.~Patil, W.~Wang, C.~Young, J.~Smith, J.~Riesa, A.~Rudnick, O.~Vinyals,
  G.~Corrado, M.~Hughes, and J.~Dean.
\newblock Google's neural machine translation system: Bridging the gap between
  human and machine translation.
\newblock {\em CoRR}, abs/1609.08144, 2016.

\end{thebibliography}
}

\end{document}